\documentclass[pdflatex,sn-mathphys,iicol]{sn-jnl}

\usepackage{
graphicx,
amsfonts,
textcomp,
program,
amssymb,
listings,
rotating,
booktabs,
algorithmicx,
mathrsfs,
url,
appendix,
etoolbox,
algpseudocode,
multirow,
algorithm,
amsmath,
hyperref,
manyfoot,
hhline,
pifont}

\usepackage{caption}
\usepackage{array}
\usepackage[export]{adjustbox}

\usepackage{multirow}
\usepackage{algpseudocode}
\usepackage{setspace}
\usepackage{subfigure}
\usepackage{bm}

\usepackage{colortbl}
\definecolor{color3}{gray}{0.95}

\jyear{2023}
\theoremstyle{thmstyleone}

\theoremstyle{thmstyletwo}

\theoremstyle{thmstylethree}

\begin{document}

\title[Article Title]{\vspace{-30pt}4DVD: Cascaded Dense-view Video Diffusion Model for High-quality 4D Content Generation}

\author[1]{\fnm{Shuzhou} \sur{Yang}}\email{\scriptsize szyang@stu.pku.edu.cn}
\author[2]{\fnm{Xiaodong} \sur{Cun}}\email{\scriptsize cun@gbu.edu.cn}
\author*[3]{\fnm{Xiaoyu} \sur{Li}}\email{xliea@connect.ust.hk}
\author[1]{\fnm{Yaowei} \sur{Li}}\email{\scriptsize liyaowei01@gmail.com}
\author*[1]{\fnm{Jian} \sur{Zhang}}\email{zhangjian.sz@pku.edu.cn}
\affil[1]{\orgdiv{Peking University Shenzhen Graduate School}, \orgaddress{\city{Shenzhen}, \country{China}}}
\affil[2]{\orgdiv{Great Bay University}, \orgaddress{\city{Dongguan}, \country{China}}}
\affil[3]{\orgdiv{Tencent}, \orgaddress{\city{Shenzhen}, \country{China}}}

\abstract{Given the high complexity of directly generating high-dimensional data such as 4D, we present 4DVD, a cascaded video diffusion model that generates 4D content in a decoupled manner. Unlike previous multi-view video methods that directly model 3D space and temporal features simultaneously with stacked cross view/temporal attention modules, 4DVD decouples this into two subtasks: coarse multi-view layout generation and structure-aware conditional generation, and effectively unifies them. Specifically, given a monocular video, 4DVD first predicts the dense view content of its layout with superior cross-view and temporal consistency. Based on the produced layout priors, a structure-aware spatio-temporal generation branch is developed, combining these coarse structural priors with the exquisite appearance content of input monocular video to generate final high-quality dense-view videos. Benefit from this, explicit 4D representation~(such as 4D Gaussian) can be optimized accurately, enabling wider practical application. To train 4DVD, we collect a dynamic 3D object dataset, called D-Objaverse, from the Objaverse benchmark and render 16 videos with 21 frames for each object. Extensive experiments demonstrate our state-of-the-art performance on both novel view synthesis and 4D generation. Our project page is \url{https://4dvd.github.io/}}


% \twocolumn[{
% \renewcommand\twocolumn[1][]{#1}
% \maketitle
% \begin{center}
%     \captionsetup{type=figure}
%     \includegraphics[width=1.\linewidth]{teaser.pdf}
%     % \captionof{figure}{Test caption}
%     \caption{\textbf{4DVD} takes a monocular video as input and generates multi-view videos. Benefiting from unprecedentedly dense-view modeling, the generated results maintain high spatial and temporal consistency.}
% \end{center}
% }]

% \begin{teaserfigure}
% \centering
%   \includegraphics[width=1\linewidth]{teaser.pdf}
%   % \vspace{-1em}
%   \caption{\textbf{4DVD} takes a monocular video as input and generates multi-view videos. Benefiting from unprecedentedly dense-view modeling, the generated results maintain high spatial and temporal consistency.}
%   \vspace{1em}
%   \label{fig:teaser}
% \end{teaserfigure}
\maketitle        % 原来的!!!!!!!!!!!!!!!!!!!!!

\begin{figure*}[t]
  \centering
    \includegraphics[width=1.\linewidth]{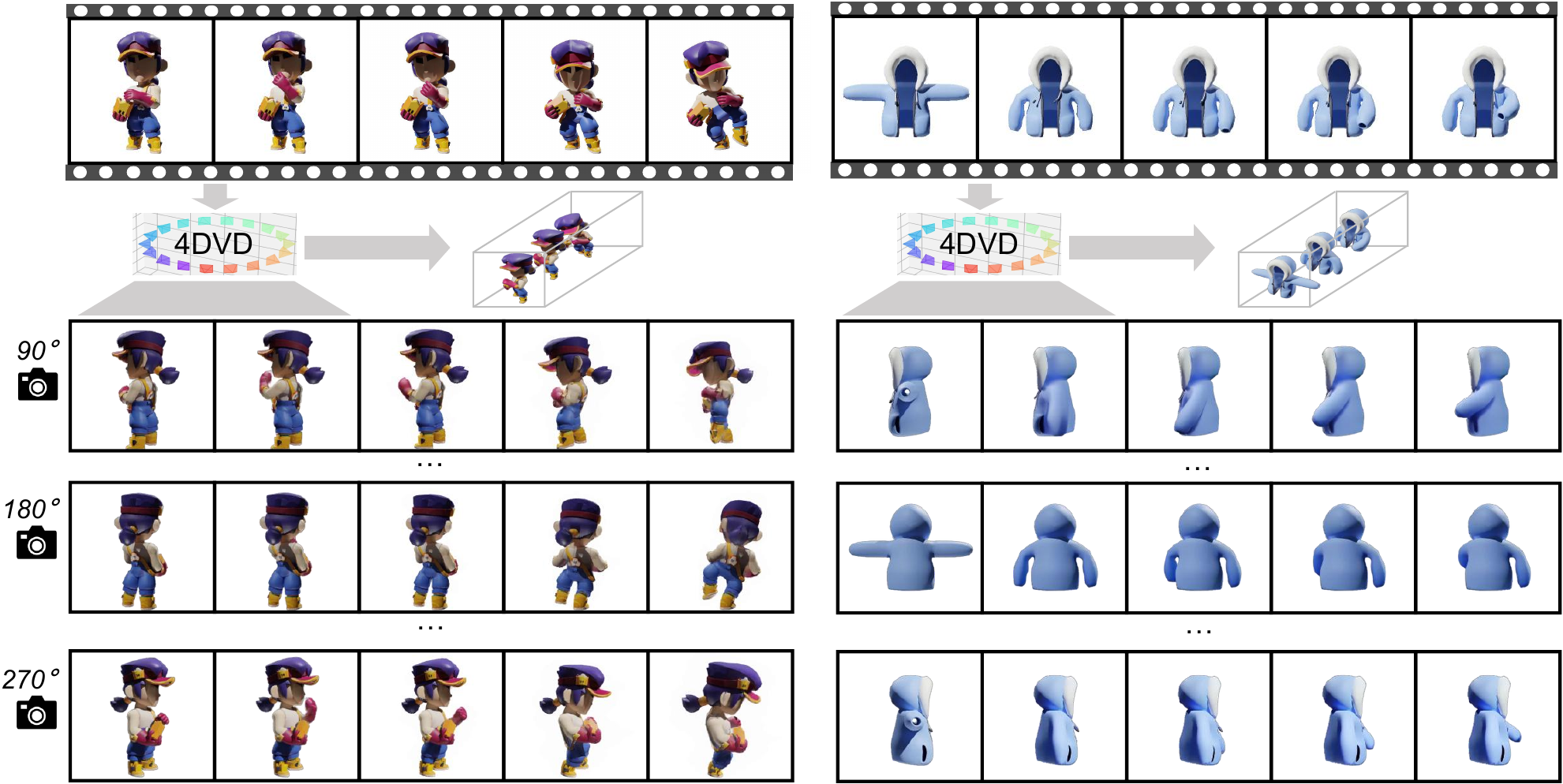}
  
  \caption{\textbf{4DVD} takes a monocular video as input and generates multi-view videos. Benefiting from unprecedentedly dense-view modeling, the generated results maintain high spatial and temporal consistency.}
  \label{fig:teaser}
  
\end{figure*}

% \vspace{-4pt}
% \let\thefootnote\relax\footnotetext{\textcolor{blue}{For reproducible research, the complete source code and training data list of the proposed 4DVD will be made publicly available when this paper is accepted.}}

\section{Introduction}
\label{sec:intro}
Dynamic 3D object generation (4D generation) aims to simultaneously create realistic 3D object content and its motion in 3D space. Considering that the 3D world we live in is inherently dynamic (such as fluttering flags, walking people, and moving objects), 4D generation is crucial to achieving immersive content creation and visual content experience. In this work, we attempt to generate 4D objects from widely available monocular videos, which enables effortless and advanced AR/VR creation.

Generating 4D asset from a given monocular video is challenging as this means the model needs to simultaneously reason about the appearance and motion around the object in 3D space based on a single 2D view, which is an ill-posed problem. To this end, recent attempts have been made to reconstruct 4D content through Score Distillation Sampling (SDS) strategy~\cite{consistent4d,Text-to-4Ddy,4Dfy,Ling_2024_CVPR,dreamgaussian4d}, which distill the 3D priors from multi-view or image generation models and temporal priors from video diffusion models. However, these methods require a time-consuming distillation process and tend to output over-smooth results with the Janus problem. Furthermore, the priors these methods distill come from different models, \textit{i.e.}, videos from video generation models, and views from image generation models, which may conflict with each other during the optimization. Another line of works~\cite{sv4d,l4gm,4Real} instead unify the prediction of multi-view and motion in a single model. Most commonly, they propose to first get the multi-view ($V$) conditions based on the first frame, then generate 4D content combined with the input monocular video ($T$). The final 4D content can be regarded as an image grid with a size of ${T \times V}$, and the model itself aims to complete the grid content based on edge conditions. Existing work is implemented through training on massive data, however, directly learning from such high-dimensional data remains difficult. On the one hand, simultaneously learning both 3D space and temporal motion with an end-to-end black-box model increases the difficulty of training. On the other hand, limited VRAM restricts the accuracy of 4D modeling. Since the model has to process $T \times V$ images in every single batch, even with the most advanced GPU, the viewpoint value $V$ can only be set to a small value (such as 4 in~\cite{4diffusion} and 8 in~\cite{sv4d}). The sparse viewpoints inevitably impact the capacity of modeling dynamic 3D space. 

To address this problem, we propose \textbf{4DVD}, a cascaded dense-view video diffusion model that aims to generate higher quality 4D assets in a decoupled manner. We find that in our case, \textit{i.e.}, multi-view video generation, spatial information is actually over-redundant, since the same visual content oftentimes appears on different views and frames. Hence, 4DVD firstly predicts consistent dense-view layouts at the expense of spatial details, and achieves structure-aware conditional generation based on these, obtaining high-quality 4D assets. Specifically, In the first stage, we downsample multi-view videos to low-resolution versions, enabling training with denser views for more efficient 3D spatial modeling. This trade-off between viewpoint number and resolution brings more prolific priors of 3D content and motion, but remains ambiguous appearance. Therefore, in the second stage, 4DVD takes multi-view layouts as the hint and operates structural conditional generation, producing high-quality multi-view results. To realize this insight, in the first stage, we inherit the structure of SV4D~\cite{sv4d}, a State-Of-The-Art (SOTA) multi-view video generation model, but extend it to denser view prediction by training on the low-resolution dense-view video data. To achieve structural-aware conditional generation of the second stage, we carefully design a unique control branch that injects multi-view layout priors to the mainstream layer-by-layer. Considering that the control branch only has multi-view layout information but lacks appearance guidance, we propose Monocular Appearance Propagation (MAP) module to further incorporate the high-quality visual content of the input monocular video in the final 4D conditional generation process. Finally, 4DVD unifies the structure modeling and structure conditional spatio-temporal generation, brings significant performance improvements and can be used for explicit 4D reconstruction. The experimental results prove the effectiveness of our proposed method. We found that open-source 4D datasets contain numerous low-quality cases, such as broken components, minor or over-excessive motion, \textit{etc}. To train 4DVD effectively, we collected a high-quality dynamic multi-view dataset from Objaverse~\cite{objaverse}, called D-Objaverse. It will be released when this paper is published.

We summarize our key contributions as follows:
\begin{itemize}
    \item[$\bullet$]
    We present 4DVD, a cascaded 4D content generation pipeline that improves generation quality by modeling 3D space and motion from unprecedented dense views.
    \item[$\bullet$]
    A structure-aware spatio-temporal generation branch is proposed to organically combine the input monocular video and dense-view coarse results, injecting these guided conditions to the mainstream model layer-by-layer and efficiently producing high-quality 4D results.
    \item[$\bullet$]
    For effective training, we carefully filter and render 16-view video data from Objaverse, called D-Objaverse. This subset contains complete object structures and natural motions.
    \item[$\bullet$]
    Both quantitative and qualitative experiments prove that 4DVD achieves state-of-the-art performance in both multiview video synthesis and 4D reconstruction.
\end{itemize}

\section{Related Work}

\subsection{3D Generation from Text or Image}
We first discuss 3D generation methods that produce 3D assets from text or images. DreamFusion~\cite{dreamfusion} proposes a Score Distillation Sampling (SDS) method to optimize the 3D content through the distillation of 2D diffusion priors, which becomes a paradigm for 3D generation using 2D diffusion models. However, due to its limited quality and time-consuming generation process, many subsequent approaches~\cite{Magic3D,ProlificDreamer,hifi123,Consistent123,gaussiandreamer,Chen_2024_CVPR,dreamcraf3d,dreamgaussian,LucidDreamer,Zhou_2024_CVPR,f123} have been proposed to develop more effective and efficient distillation methods, aiming to address the Janus problem, alleviate over-smoothed content, and speed up the generation process. 
To overcome the time-consuming optimization process of these SDS-based methods for each generated 3D object, researchers have explored direct prediction of 3D models in a feedforward manner via a large reconstruction model~\cite{LRM,LEAP,pflrm,Zou_2024_CVPR,MeshLRM,triposr,LGM}, which could generate the 3D model instantly. However, due to the high complexity of the 3D representations such as NeRF~\cite{NeRF}, 3D Gaussians~\cite{3dgs}, and triplane and limited network capacity, these methods usually could only generate objects with simple geometry and texture. 
Other works propose multi-view generation models~\cite{mvdream,ImageDream,Wonder3D,Era3D} that can produce view-consistent multi-view images of the 3D content. Compared with direct 3D generation via a large reconstruction model, reconstructing 3D content using the generated multi-view images greatly reduces the complexity of directly learning 3D representations and achieves superior performance. Inspired by this strategy, we develop a new algorithm that produces consistent multi-view videos (instead of images) to reconstruct 4D objects.

%However, due to the lack of a unified 3D representation, different algorithms predict different 3D models, such as mesh, 3D Gaussian~\cite{3dgs}, and NeRF~\cite{NeRF}, increasing the difficulty of practical application. On the other hand, compared with image generation, directly training models to predict 3D parameters is more complex, which limits the generation quality. 

\begin{figure*}[t]
  \centering
    \includegraphics[width=1.\linewidth]{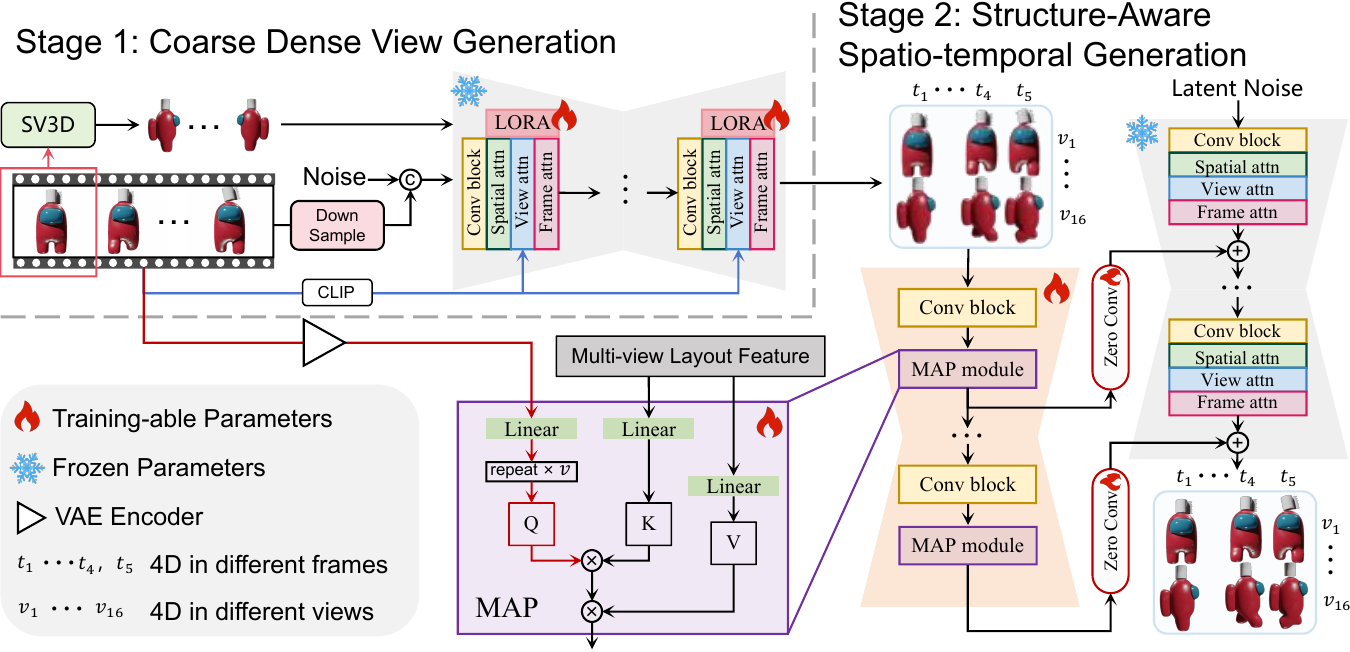}
  
  \caption{\textbf{Pipeline of 4DVD}. Our model consists of two stages: The coarse dense view generation receives a monocular video and reference views, producing the coarse 16-view layouts. Based on layout condition, the structure-aware spatio-temporal generation predicts detailed multi-view videos. To get appearance guidance, we take input monocular video as an external condition and develop Monocular Appearance Propagation (MAP) module to integrate it, as shown in the purple region.}
  \label{fig:workflow}
  % \vspace{-0.5em}
  
\end{figure*}

\subsection{4D Content Generation}
The development of object-centered 4D content generation is similar to that of 3D generation. Optimization-based methods are also firstly explored~\cite{consistent4d,Text-to-4Ddy,4Dfy,Ling_2024_CVPR,dreamgaussian4d,animate124,4dgen,STAG4D, efficient4d, diffusion2}, using SDS loss to optimize the 4D representation~\cite{D-NeRF,HexPlane,4dgs} through pre-trained video diffusion models. The key insight of these methods is the distillation of a text-to-video diffusion model and a multi-view diffusion model to get temporal and cross-view priors. For example, 4D-fy~\cite{4Dfy} divides the optimization process into three stages, training based on the multi-view model, image model, and video model, which improves 3D consistency, generation quality, and motion quality respectively. However, these methods can take hours to produce 4D content due to the time-consuming distillation process. In addition, large reconstruction models~\cite{l4gm} and multi-view video generation models~\cite{sv4d,4diffusion,diffusion4d} are also developed. The former directly predicts 3D Gaussian parameters for each frame, which requires enormous GPU resources for training, and its generation quality is limited by complex modeling of implicit prediction parameters. The latter predicts 3D consistency multi-view videos, which require fewer computing resources and enable high-quality generation. However, due to the limited VRAM, current multi-view video generation methods can only model dynamic 3D space with relatively sparse views. This not only restricts the ability of generation models to perceive the full 3D space but also cannot be used to reconstruct explicit 4D representations such as 4D Gaussian accurately, which usually requires dense views as input. More recently, another line of works~\cite{DimensionX, 4Real, TrajectoryCrafter} attempted to generate 4D scenes containing multiple objects and wide background. But these methods mainly focus on extending monocular scene videos to flexible forward views for immersive watching experience, instead of facilitating dynamic 3D digital asset creation. To produce full 360-degree views for object-centered creation, we propose a cascaded dense-view video generation model to produce highly cross-view consistent multi-view videos with high-quality details.

\section{Method}
\label{sec:method}
Given a monocular video $\textbf{I}\in \mathbb{R}^{T \times D}$ of a dynamic object with $T$ frames and $D = 3\times H\times W$ dimensions, 4DVD aims to faithfully predict the dynamic content in different views with both view and temporal consistency. Specifically, our goal is to extend the video $\textbf{I}$ from $\mathbb{R}^{T \times D}$ to $\mathbb{R}^{T \times V \times D}$, where $V$ is the number of viewpoints. Considering more viewpoints facilitate more complete modeling for dynamic 3D spaces, in this work, we introduce a coarse-to-fine generation pipeline that produces highly consistent dense-view videos in a decoupled-cascaded manner.

\subsection{Preliminaries: Latent Diffusion Models} Existing multi-view video diffusion models can be regarded as a kind of Latent Diffusion Model (LDM). These models~\cite{sd,wan} generate visual content in the latent space that can be encoded and decoded by the Variational Auto-Encoder (VAE) model. In training, given the latent $\textbf{z}_{0}$ encoded by the VAE encoder, we add noise ${\epsilon}$ of various scales to it following a predefined schedule defined as:
\begin{equation}
\label{eq:diff_1}
\mathbf{z}_t = \sqrt{\bar{\alpha}_t}\mathbf{z}_0 + \sqrt{1-\bar{\alpha}_t}\mathbf{\epsilon},
\end{equation}
LDM is trained to precisely predict the added noise ${\epsilon}$ in the noisy latent $\mathbf{z}_{t}$ with a neural network $\epsilon_{\theta}(\cdot)$, whose objective function can be summarized as:
\begin{eqnarray}
\begin{aligned}
    \label{sd}
    \underset{\boldsymbol{\theta}}{\min} \, \mathbb{E}_{\textbf{z}_0, \boldsymbol{\varepsilon} \sim \mathcal{N}(0,I), t} \|\boldsymbol{\varepsilon} - \varepsilon_{\boldsymbol{\theta}}(\textbf{z}_t, t, \textbf{c})\|_2^2,
\end{aligned}
\end{eqnarray}
where $t$ is the sampling step and $\mathbf{c}$ is the text condition guiding the generated content. After training, $\epsilon_{\theta}(\cdot)$ receives a random latent noise $\textbf{z}_T$, denoises step by step with a prompt condition $\textbf{c}$, and decodes the final latent through a VAE decoder to get final results.

\subsection{Cascaded Coarse-to-fine Pipeline}
Simultaneously generating $T \times V$ images requires massive VRAM and is inefficient. Previous methods that process the $T \times V$ image grid usually restrict to limited viewpoints, which makes it challenging to fully comprehend the geometry and motion in 3D space and reconstruct the 4D representation accurately, as this requires highly consistent dense-view input. To fully unleash the potentiality of generation models to create dynamic content in 3D space, we propose a two-stage cascaded pipeline as shown in Fig.~\ref{fig:workflow} to generate prolific multi-view videos from a monocular input video. Our key idea is to learn dense-view layout priors to model the primitive structure of 4D objects precisely in the coarse stage. And then, guided by these coarse structural priors, we synthesize high-quality final results while maintaining the view and temporal consistency in the fine stage. Note that since spatial information is redundant across view and time, resolution can actually be sacrificed to learn multi-view layout priors, allowing for more adequate and accurate 3D modeling from more viewpoints. These dense views bring more consistent generation results for 3D content and object motion. In the fine stage, we dynamically select a subset of the dense views in each training iteration to supervise under limited VRAM. Considering that the fine stage aims to perform spatio-temporal generation guided by the layout hints from the coarse stage, instead of purely generate the dynamic 3D space from scratch, it is sufficient and effective to optimize with only a part of viewpoints in each iteration. During inference, due to the absence of back propagation, 4DVD can output the complete multi-view videos based on the full 16-view layout guidance in the fine stage. This coarse-to-fine process aligns with the workflow of the human visual system: we first perceive the general layout of a dynamic object at a glance and then focus on enriching its details.

% In the fine stage, we dynamically select a subset of the dense views in each training iteration to supervise in order to save GPU memory. Considering that the fine stage aims to perform detail enhancement guided by the 4D results from the coarse stage instead of modeling the dynamic 3D space from scratch, it is sufficient and effective to optimize with only a part of the views in each iteration. This coarse-to-fine process aligns with the workflow of the human visual system: we first perceive the general layout of a dynamic object at a glance and then focus on enriching its details.

\begin{figure}[t]
  \centering
    \includegraphics[width=1.\linewidth]{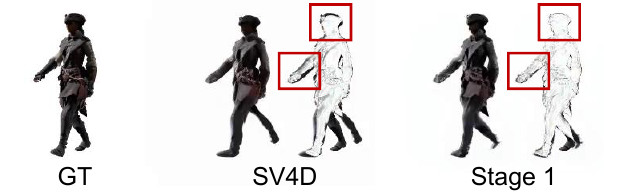}
  \vspace{-1.em}
  
  \caption{\textbf{Visual and residual results}. We provide the GT image, results of SV4D and our first stage model, and the corresponding residual images compared to the GT image. Enforcing SV4D to produce 16 views leads to large errors (as shown in regions boxed in the red), while results from our first stage achieve better results after fine-tuning.}
  \label{fig:fig2}
  % \vspace{-1.em}  这里开放就能省很多空间!!!!!!!!!!!!!!!!
\end{figure}

\subsection{Coarse Dense View Generation}
Previous multi-view video generation methods like SV4D~\cite{sv4d,sv4d2} are trained on sparse view videos, which could also be adapted to predict more views by directly adjusting their hyperparameters during inference. But this naive modification has some problems. First, it fails to predict accurate results when the number of viewpoints is greater than the number of views in the training set. As illustrated in Fig.~\ref{fig:fig2}, we present visual results produced by SV4D with 16 viewpoints and their Ground Truth (GT) and visualize their residual image on the right. One can see that although SV4D could generate more views, its output exhibits abnormal geometry such as the distorted right leg, and significantly deviates from the GT image. Highlighted in red boxes of Fig.~\ref{fig:fig2}, the contours of the residual image exhibit obvious differences, indicating that the overall output layout of SV4D does not accurately match the GT view. This is because SV4D is restricted by the sparse training views~(8 views on the most advanced GPU), making it challenging to fully model the 4D space. Forcing generalization and producing more views exposes this inability to accurately depict 4D content.

To address this problem, we propose to generate denser view videos at the expense of spatial content. We finetune SV4D on dense view low-resolution videos (e.g., 16 views in our setting) using LORA~\cite{lora} to fit the VRAM requirements. Our goal is not to directly predict the 16-views spatio-temporal information, but only to obtain the 4D layout priors that are consistent across views. As shown in Fig.~\ref{fig:workflow}, given a monocular video as input, we initially employ the existing image-to-3d method \cite{sv3d} to synthesize novel views for the first frame. Taking the reference video and 16 reference views as input, the network in the coarse stage predicts the $T \times V$ image grid at a low resolution. Considering spatial messages are actually over-redundant in our case, we believe that sacrificing it in exchange for a more adequate modeling of 4D space is worthwhile and effective. We incorporate the LORA module into three attention blocks, which are used to process spatial features, fuse with the input video, and fuse with the reference views, respectively. 

% We introduce LORA for fine-tuning to adapt the model to generate consistent dense view results. 
As shown in Fig.~\ref{fig:fig2}, the output of Stage 1 exhibits reliable geometry, and its residual image exhibits less difference, demonstrating that the unique design in our coarse stage leads to more accurate 4D modeling and generation. Theoretically, we could use even more viewpoints for training by further reducing video resolution. However, we observe that the VAE encoder-decoder \cite{sd} cannot reproduce images with too low resolution faithfully. We believe this is because it is pre-trained on high-quality images with high resolutions. Consider that 16 views already cover 4D object densely enough, and further reducing spatial messages has the potential risk of even erasing desired layout information. According to our experiments, we choose $256 \times 256$ as the final resolution for our first stage, which is enough to enable dense-view training.

\subsection{Structure-Aware Spatio-temporal Generation}
Since we produce multi-view videos simultaneously with attention mechanisms, they preserve consistent structure information spatially and temporally. However, these layouts exhibit blurred textures, low resolution, and quality that deviate from the input video. Therefore, our goal in the fine stage is to produce high-quality dense view videos based on these consistent layouts, and we introduce structure-aware spatio-temporal generation to achieve this.

Inspired by previous conditional generation methods~\cite{controlnet, t2iadapter, animateanyone}, which extract layout features through a specific branch and inject them to the base model, we also utilize SV4D as mainstream model and introduce a trainable structure condition branch. It encodes the coarse multi-view layouts to latent space that is aligned to mainstream model, and injects them into the base model. As shown in Fig.~\ref{fig:workflow}, we take the output of the first stage as the input condition of our structure condition branch, which is stacked with convolution blocks and our proposed MAP modules. During training, we freeze the base model and update the parameters of the structure condition branch only. Constrained by the limited VRAM, we randomly select 4 views of the $576 \times 576$ resolution at each training iteration. Since the condition branch injects the aligned layout features into the base model, and the goal of the second stage is not to model 4D structure but to provide better layout guidance for each view, there is no need to provide full-view condition.

Note that this stage aims to generate multi-view videos with elegant appearance, so high-quality appearance prior is required but so far only multi-view layout conditions are available. Therefore, we utilize the input monocular video as external condition and propose Monocular Appearance Propagation (MAP) module to propagate its appearance to multiple views. As shown in Fig.~\ref{fig:workflow}, in each MAP module, the high-quality input video is injected to provide appearance reference. The details of the MAP module are given in the purple region, which is a basic cross-attention design with some unique modifications. Note that conventional cross-attention uses the features of the current model to calculate the \textbf{Q} matrix and employs external conditions for \textbf{KV} matrices. But in our case, we calculate \textbf{KV} matrices with the multi-view features and use reference video for \textbf{Q} matrix. This is because our condition branch features structure information of multi-view videos whose shape is $T \times V$, but the reference video only records the high-quality content of a single view ($T \times 1$). Therefore, MAP module actually aims to integrate these two unaligned features. Focusing on this special issue, we swap the calculation of \textbf{Q} and \textbf{KV}, enabling the model to extend high-quality reference video to novel views based on the condition of multi-view layouts, which is easier to train than conventional operations that forcibly upscale the multi-view coarse results based on monocular video. Experiments in Sec.~\ref{subsec:abla} prove the effectiveness of our unique design.

To enable this modified cross-view calculation, we repeat the reference video for $V$ times before multiplication. Denote the linear layers that calculate \textbf{Q}, \textbf{K}, \textbf{V} as $L_Q$, $L_K$, $L_V$ respectively, the operation of MAP can be expressed as:
\begin{eqnarray}
\begin{aligned}
    \label{shi1}
    %F^{ref} &= Repeat(Resize(F^{ref}, WH), V) \in \mathbb{R}^{WHV \times 4}, \\
    \textbf{Q} &= L_Q(F^{ref}) \in \mathbb{R}^{WHV \times N_c}, \\
    \textbf{K} &= L_K(F^{LR}) \in \mathbb{R}^{N_c \times WHV}, \\
    \textbf{V} &= L_V(F^{LR}) \in \mathbb{R}^{WHV \times N_c}, \\
\end{aligned}
\end{eqnarray}
where $WH$ and $N_c$ mean the spatial resolution and feature channel number of the current layer, $F^{ref}$ is the normalized VAE feature of the reference video, and $F^{LR}$ means the coarse multi-view layout features. For training, we set $V=4$ and during inference, we use the full coarse image grid as the condition, \textit{i.e.}, $V=16$. 

%The design of RSR attention lies in the unique unaligned problem in our case. In practice, we find using conventional cross-attention is challenging for the network to upscale full-view coarse results based on the single-view condition. 

%Experiments in Sec.~\ref{subsec:abla} demonstrate this issue. In contrast, swapping the calculation of \textbf{Q} and \textbf{KV} means enforcing the model to modify the high-quality reference video with the condition of full-view coarse results, which is relatively easy to train. Some video super-resolution methods \cite{Kim_2023_WACV} also adopt a similar design.

\subsection{Training}
We train our cascaded model in a decoupled manner for efficiency. In the first stage, we enforce the model to produce 16-view videos with the conditions of reference video $\textbf{c}_f$ and view $\textbf{c}_v$, objective function can be expressed as:
\begin{eqnarray}
\begin{aligned}
    \label{shi2}
    \underset{\boldsymbol{\theta}}{\min} \, \mathbb{E}_{\textbf{z}_0, \boldsymbol{\varepsilon} \sim \mathcal{N}(0,I), t} \|\boldsymbol{\varepsilon} - \varepsilon_{\boldsymbol{\theta}}(\textbf{z}_t, t, \textbf{c}_f, \textbf{c}_v)\|_2^2,
\end{aligned}
\end{eqnarray}
where $\theta$ is the parameters of our added LORA. In the second stage, we optimize the condition branch to realize structure-aware spatio-temporal generation, which is conditioned on the randomly sampled 4 coarse layouts $\textbf{I} \in \mathbb{R}^{T \times 4}$. Using $\theta$ to represent trainable parameters and loss function is formulated as:
\begin{eqnarray}
\begin{aligned}
    \label{shi3}
    \underset{\boldsymbol{\theta}}{\min} \, \mathbb{E}_{\textbf{z}_0, \boldsymbol{\varepsilon} \sim \mathcal{N}(0,I), t} \|\boldsymbol{\varepsilon} - \varepsilon_{\boldsymbol{\theta}}(\textbf{z}_t, t, \textbf{I})\|_2^2,
\end{aligned}
\end{eqnarray}

\begin{figure*}[t]
  \centering
    \includegraphics[width=1.\linewidth]{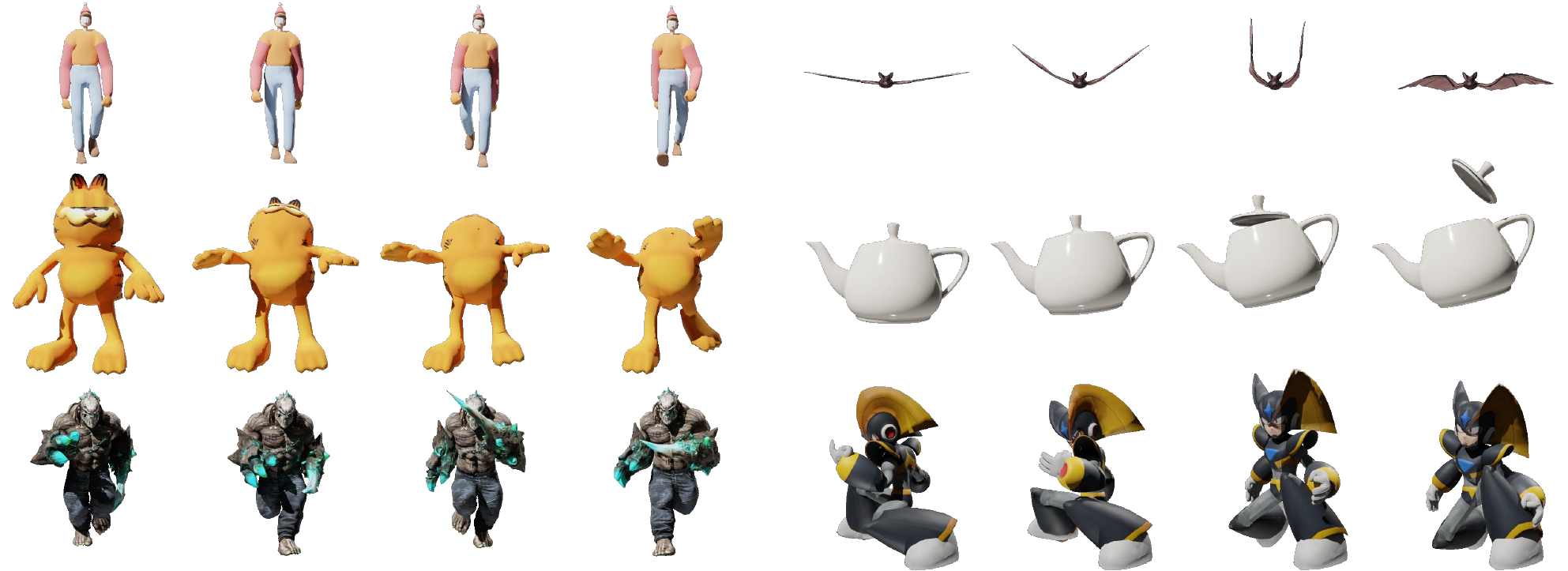}
  \vspace{-1.em}
  
  \caption{\textbf{Some cases in D-Objaverse}. We carefully filtered and rendered a high-quality dynamic 3D datasets from Objaverse data, called D-Objaverse. Here we showcase some front-view cases of D-Objaverse.}
  \label{fig:supp_data}
  
\end{figure*}

\begin{figure*}[t]
  \centering
    \includegraphics[width=1.\linewidth]{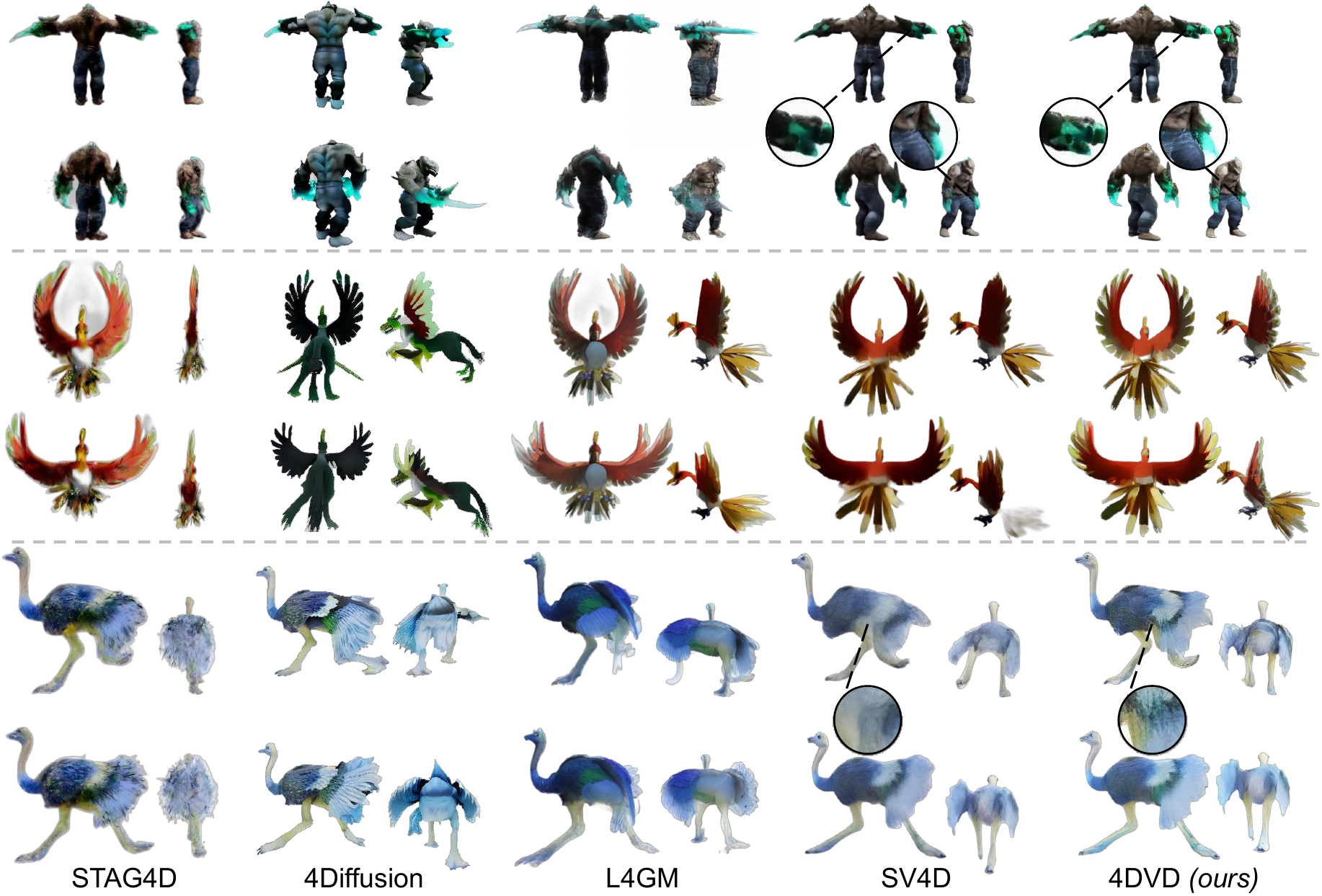}
  \vspace{-1.em}
  
  \caption{\textbf{Visual comparison of the generated multi-view videos}. We show two timesteps and two views for each example. Compared to other baseline methods, 4DVD enables higher quality generation which has finer textures, reasonable appearance, and better consistency.}
  \label{fig:com_video}
  \vspace{-0.5em}
  
\end{figure*}

\section{D-Objaverse Data}
Although Objaverse contains numerous 3D objects, most of them are static and cannot be used to train multi-view video models. Some open-source dynamic 3D data is available such as ObjaverseDy used by SV4D or Consistent4D. However, we notice that these available dynamic 3D data either only contain limited examples or have quite a lot of low-quality cases. These low qualities can be summarized into the following three manifestations: \textbf{1)} Incomplete structure. Many 3D objects are just parts of common objects, such as an arm or a leg. \textbf{2)} Abnormal motion. Some cases exhibit excessive or unnoticeable movement. \textbf{3)} Undefined content. Some examples show blurry/chaotic content, although they are also dynamic 3D objects. To effectively train multi-view video model, we filter out high-quality dynamic 3D assets and render high-resolution videos in 16 viewpoints. We have designed three filtering indicators specifically for the three main low-quality situations. \textbf{1)} We use LPIPS to select samples of acceptable quality. \textbf{2)} We filter out samples with reasonable motion by calculating the optical flow between frames. \textbf{3)} And at last, we utilize a multimodal LLM (\textit{i.e.}, CogVLM~\cite{CogVLM}) to filter out examples containing incomplete structures. We collected 41k dynamic 3D assets based on ObjaverseDy, containing 27k samples, and other accessible dynamic 3D data, and finally filtered out 17k samples with 16-view videos that met our criteria. This high-quality dnamic 3D dataset is called D-Objaverse since it is mainly filtered from Objaverse. As a major contribution of 4DVD, D-Objaverse will be released when 4DVD is published. We showcase some frames of front-view videos from D-objaverse in Fig.~\ref{fig:supp_data}.

\section{Experiments}

\begin{table*}
\centering
  \caption{\textbf{Quantitative comparison of produced multi-view videos and produced 4D Assets.} 4DVD achieves superior performance in visual quality, video frame consistency and multi-view consistency. The best results are highlighted in \textbf{bold}.}
  % \vspace{-1.em}
  \label{tab:com_video}
  \resizebox{\linewidth}{!}{
  \begin{tabular}{l|l|ccccc|ccc}
    \toprule
    \multirow{2}{*}{Model} & \multirow{2}{*}{Type}  & \multicolumn{5}{c|}{Multi-View Videos} & \multicolumn{3}{c}{4D Assets}  \\ \cline{3-10}
    &  & LPIPS $\downarrow$ & CLIP-S $\uparrow$ & FVD-F $\downarrow$ & FVD-V $\downarrow$ & FVD-Diag $\downarrow$ & LPIPS $\downarrow$ & CLIP-S $\uparrow$ & FVD $\downarrow$ \\  \hline
    STAG4D & Optimization-based & 0.156 & 0.892 & 714.51 & 528.06 & 652.17 & 0.157 & 0.902 &  879.19  \\
    L4GM & Large reconstruction model & 0.144 & 0.910 & 606.90 & 467.81 & 525.22 & 0.143 & 0.911 & 543.57 \\
    4Diffusion & Multi-view video model & 0.155 & 0.899 & 853.89 & 776.92 & 816.88 & 0.165 & 0.874 & 822.29 \\
    SV4D & Multi-view video model & 0.136 & 0.913 & 586.41 & 482.64 & 542.87 & 0.161 & 0.897 & 677.56 \\ \hline
    4DVD & Multi-view video model & \textbf{0.133} & \textbf{0.927} & \textbf{507.12} & \textbf{314.44} & \textbf{456.01} & \textbf{0.136} & \textbf{0.919} &   \textbf{438.41}\\
    \bottomrule
  \end{tabular}
  }
  % \vspace{-1.5em}
\end{table*}

\subsection{Implementation Details}
\paragraph{Hyperparameter Settings.} We train and evaluate the performance of different methods on the collected D-Objaverse dataset. For a fair comparison, we randomly select 50 dynamic objects for evaluation and the remaining ones are used for training. We use Blender to render 16-view videos with the resolution of $576 \times 576$ for each 4D asset. Each video clip is 10 FPS and is downsampled to a corresponding low-resolution version (\textit{i.e.}, $256 \times 256$) for the coarse stage. During training, we set $\textit{T}=5$. For the first stage, we supervise the network with 16-view low-resolution videos. For the second stage, we randomly select 4 views from the high-resolution videos and make the model to produce a $5 \times 4$ high-resolution image grid in each iteration, where the corresponding 4 views of low-resolution version are adopted as structural generation conditions. During inference, the model can predict the entire high-resolution $5 \times 16$ image grid at once within 30GB VRAM. To extend to 21-frames full videos, we follow the same anchor-sampling approach as SV4D. D-Objaverse will be released when this paper is published.

\paragraph{Metrics.} Following previous methods~\cite{sv4d,l4gm}, we evaluate the generation quality of different methods with three widely used metrics: Learned Perceptual Similarity~\cite{LPIPS} ($LPIPS$), CLIP-score ($CLIP\text{-}S$), and $FVD$. Considering the characteristics of $T \times V$ 4D generation, SV4D~\cite{sv4d} proposes some variants of $FVD$ and we adopt three of them, 1) $FVD\text{-}F$: fix the viewpoint and calculate $FVD$ across frames. 2) $FVD\text{-}V$: fix the frame and calculate $FVD$ across views. 3) $FVD\text{-}Diag$: calculate $FVD$ over the diagonal images of the $T \times V$ image grid. These three metrics could evaluate the 4D content in terms of time, space, and space-time, respectively.

\paragraph{Baselines.} We compare our method with other four well-known related methods for 4D generation from a monocular video, including SDS-based method STAG4D~\cite{STAG4D}, multi-view video generation models 4Diffusion~\cite{4diffusion}, SV4D~\cite{sv4d}, and dynamic large reconstruction model L4GM~\cite{l4gm}.

\begin{figure*}[t]
  \centering
    \includegraphics[width=1.\linewidth]{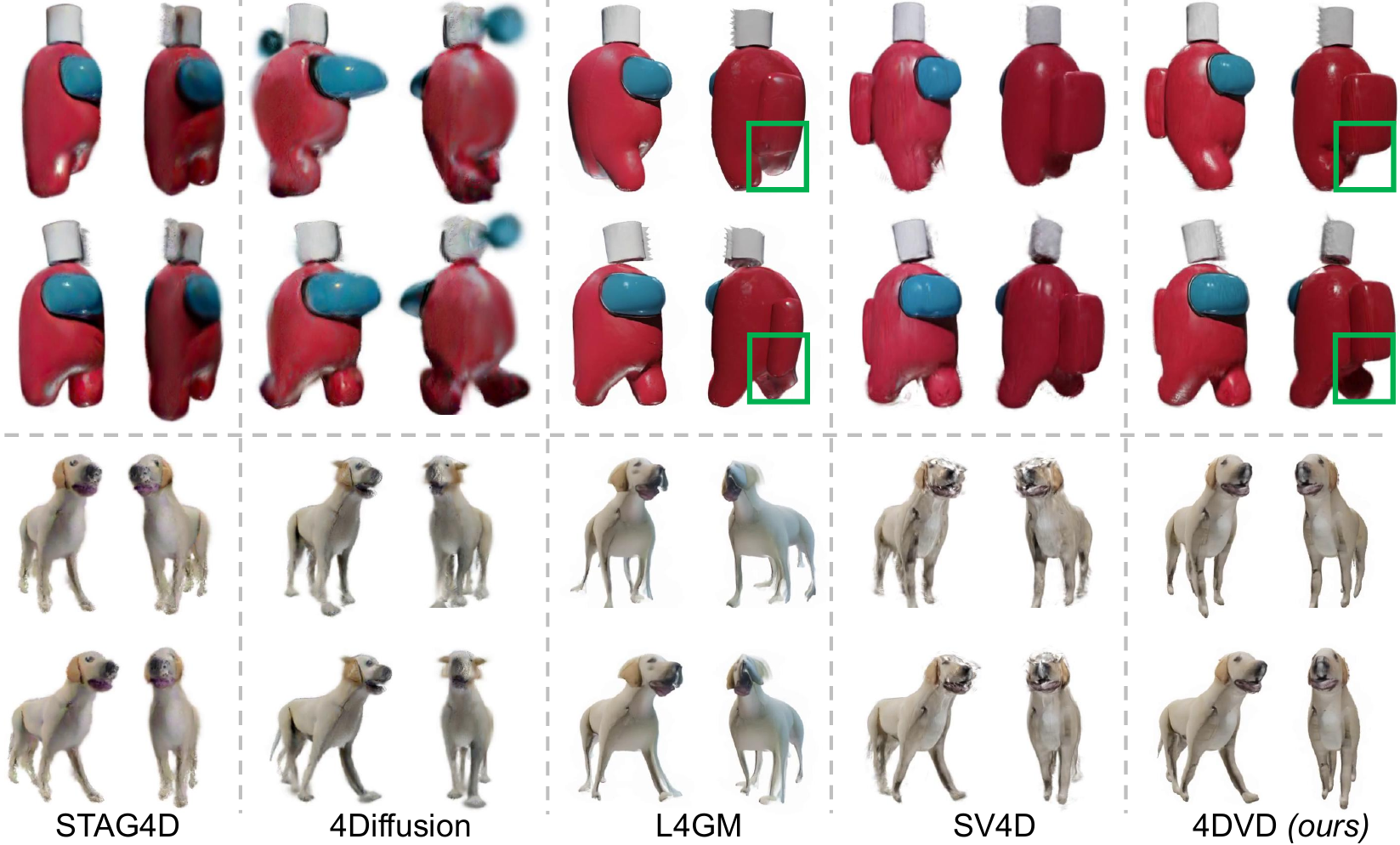}
  \vspace{-1.em}
  
  \caption{\textbf{Visual comparison of the generated 4D outputs}. We show two timesteps and two views for each example. By leveraging the dense-view videos generated by our method, 4D Gaussian can be reconstructed well with high-quality geometry and textures. Compared to prior works, our results are more consistent and detailed with a more reasonable 3D structure and satisfactory appearance.}
  \label{fig:com_4d}
  
\end{figure*}

\begin{figure*}[t]
  \centering
    \includegraphics[width=1.\linewidth]{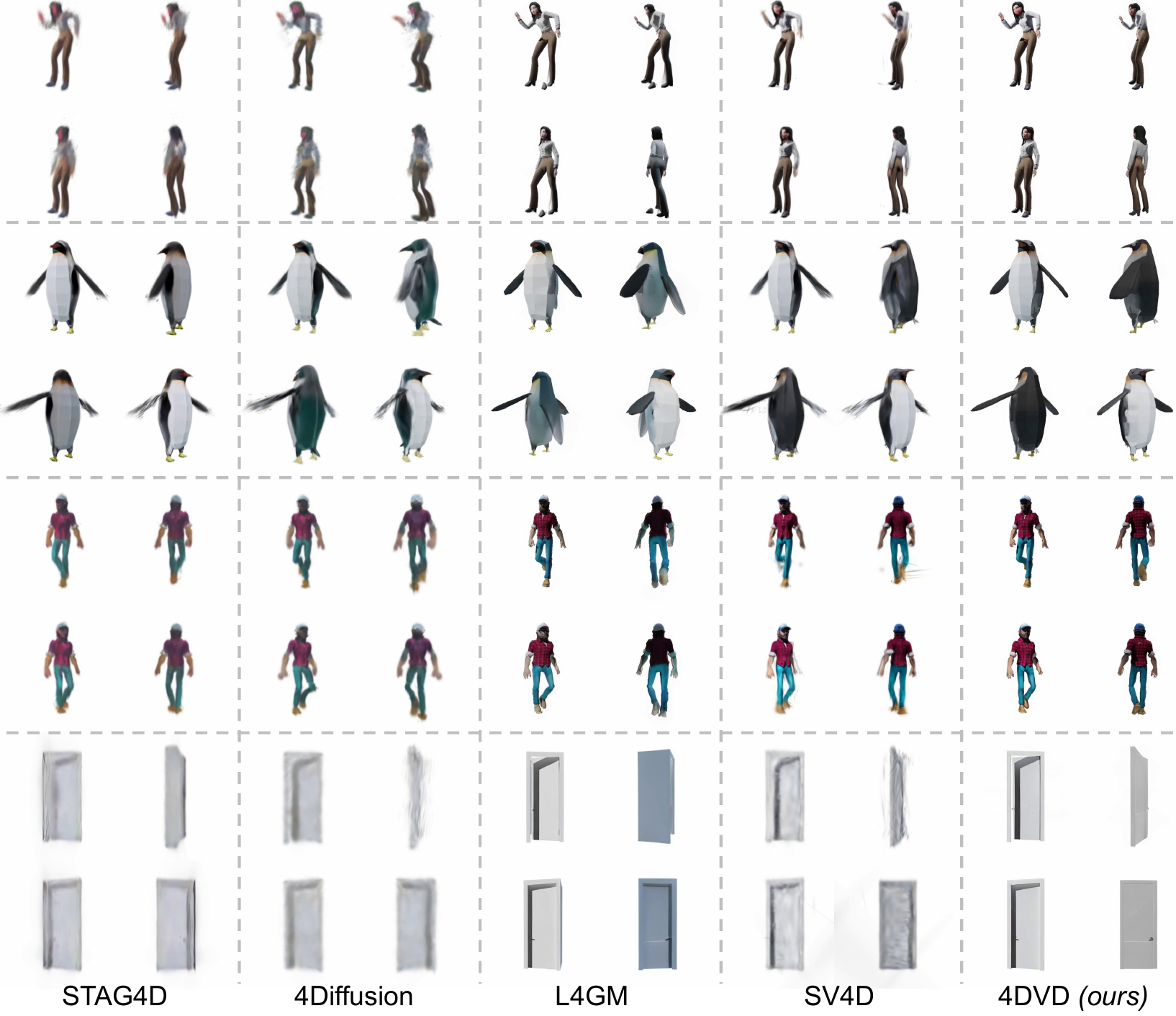}
  \vspace{-1.em}
  
  \caption{\textbf{More visual comparison of the generated 4D outputs from different methods}. Similiar to Fig.~\ref{fig:com_4d}, we show two time steps and two views for each example. Compared to prior works, our results exhibit clearer appearance and more prolific geometric structures. We believe this performance advantage mainly comes from the more consistent multi-view video and unprecedented dense viewing angles produced by 4DVD.}
  \label{fig:supp_com4d}
  
\end{figure*}

\subsection{Quantitative Comparison}

\paragraph{Multi-view Video Synthesis.} Considering that our method can achieve 16-view generation but others cannot, we only use videos from the four orthogonal viewpoints for comparison. As shown in Tab.~\ref{tab:com_video}, we calculate evaluation metrics on the $T (21) \times V (4)$ image grids produced by different methods. For STAG4D and L4GM, we render the same 4 views from their outputs to get image grids. One can see that even when only compared to the orthogonal 4-view videos, 4DVD still achieves better results on both cross-view consistency and visual quality.

\paragraph{4D Generation.} To compare the quality of 4D assets, we use the same 4D Gaussian representation following DreamGaussian4D~\cite{dreamgaussian4d} to reconstruct for multi-view video methods. For STAG4D~\cite{STAG4D} and L4GM~\cite{l4gm}, we directly use their outputs. After getting the 4D Gaussian of different methods, we could render videos with both camera view and temporal index change simultaneous for comparison in 4D. The details of rendered free-view videos can be found in supplementary materials. As we evaluate the free-view video instead of the image grid, we could directly use $FVD$ to measure spatio-temporal quality. We report the quantitaive comparison between our results and the baselines in Tab.~\ref{tab:com_video}. Our method performs best in terms of all metrics, demonstrating its superiority in visual quality ($LPIPS$, $CLIP\text{-}S$), 3D and temporal consistency ($FVD$). Although 4Diffusion and SV4D can produce high-quality videos, their limited sparse views lead to visual degradation of their 4D Gaussian results. In contrast, our method enables 16-view generation, which not only brings more consistent results, but also allows high-quality 4D reconstruction. One can see that among all SOTA methods, 4D assets of 4DVD achieves the best scores across all metrics.

\subsection{Visual Comparison}

\paragraph{Multi-view Video Synthesis.} In Fig.~\ref{fig:com_video}, we showcase the video results produced by different methods on three examples with two different angles and timesteps. Notice that our method enables 16-view generation and other methods cannot, for comparison, the selected two viewpoints are orthogonal, which can be directly produced by all approaches. We observe that the SDS-based methods (\textit{e.g.}, STAG4D) fail to produce reasonable 3D geometries and suffer from the Janus problem. Take the example in the middle of Fig.~\ref{fig:com_video}, where we show the results in back and side views, but the back view of STAG4D still has the frontal head, and the side view is only a slice. 4Diffusion, which is a multi-view video model, enables multi-view generation while maintaining consistency across views and frames. However, its results exhibit weird structures and unnatural appearance. The most recent L4GM and SV4D achieve relatively higher quality, but L4GM cannot predict geometry well. We believe this is because directly predicting 3D Gaussian parameters for all frames is complicated, which causes L4GM to be unable to accurately predict the novel view content of cases given in Fig.~\ref{fig:com_video}. Meanwhile, SV4D tends to produce over-smooth texture, leading to unsatisfactory appearances. In contrast, benefiting from the proposed decoupled generation strategy, our cascaded model simplifies the difficulty of modeling 3D content and motion, producing better visual results.

\paragraph{4D Generation.} For the 4D reconstruction results, as shown in Fig.~\ref{fig:com_4d}, STAG4D still suffers from the Janus problem and fails to model the motion in 3D space well. 4Diffusion mistakes the 3D content and produces distorted structures. For the cases such as the dog with slender legs, L4GM fails to model it since this model generates Gaussian parameters based on only 4 views, which is sparse for 4D modeling. SV4D only produces 8-view videos, which is not enough to reconstruct the explicit 4D representation. In contrast, our method that directly generates 16 consistent dense view videos enables more accurate and consistent 4D modeling, which can be used to reconstruct 4D Gaussians with high quality. Moreover, to provide stronger evidence, we further display more comparison cases in Fig.~\ref{fig:supp_com4d}. One can see that previous work cannot produce high-quality 4D assets while 4DVD enables satisfactory 4D generation. We believe this is mainly due to two reasons. First, due to the advanced novel workflow, 4DVD can produce multi-view videos that are highly consistent in time and viewpoint, which enables accurate 4D reconstruction. Second, 4DVD produces unprecedented dense views, providing ample supervision for 4D modeling.

\begin{figure*}[t]
  \centering
    \includegraphics[width=0.9\linewidth]{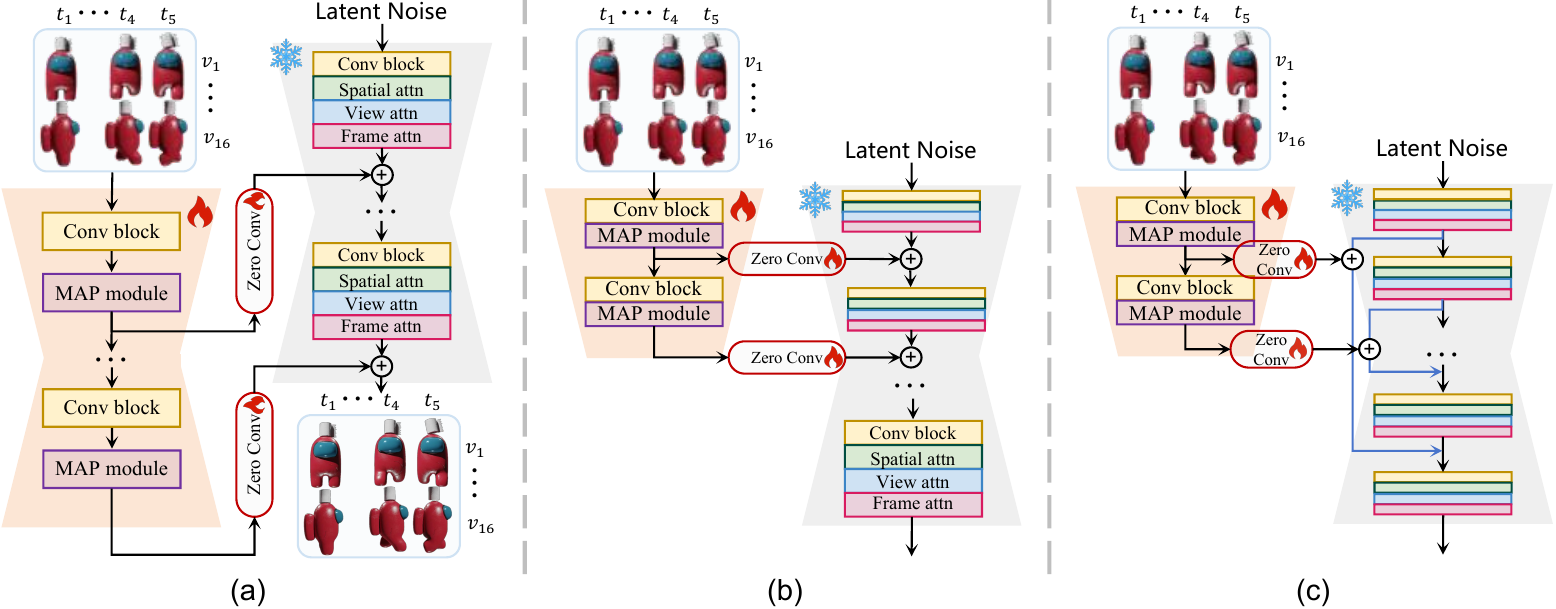}
  
  \caption{\textbf{Network architectures of different condition branches}. We show three optional structure condition branches used in our second stage: (a) our method, (b) T2I-adapter, and (c) ControlNet.}
  \label{fig:supp_workflow}
  
\end{figure*}

\begin{figure*}[t]
  \centering
    \includegraphics[width=1.\linewidth]{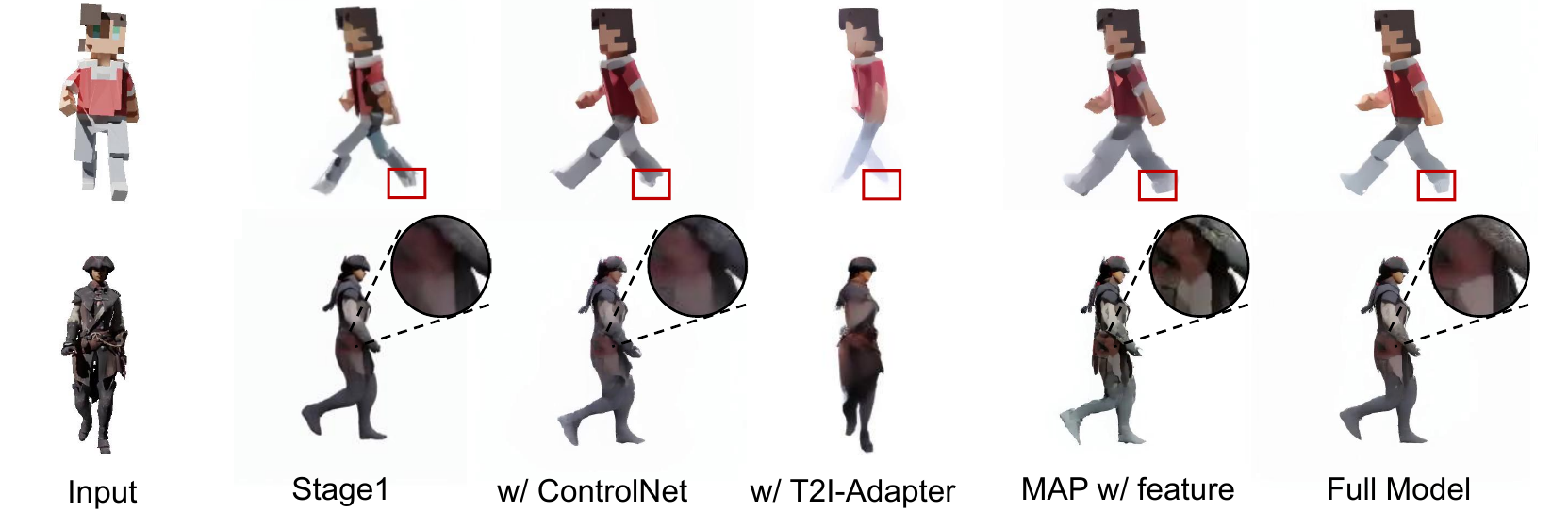}
  \vspace{-1.em}
  
  \caption{\textbf{Visual results of ablation study}. Results from stage 1 are rough and only serve as a hint for subsequent generation. Compared with mainstream condition braches such as ControlNet or T2i-adapter, the branch we developed performs best. We also adapt the MAP module to take the model feature as the \textbf{Q} matrix, which introduces undesired color offset and artifacts.}
  \label{fig:supp_abla}
  % \vspace{-0.5em}
  
\end{figure*}

\begin{table}
\centering
  \caption{\textbf{Comparison on runtime.} The best results are highlighted in \textbf{bold}.}
  \label{tab:runtime}
  \resizebox{\linewidth}{!}{
  \begin{tabular}{l|ccccc}
    \toprule
    Methods & STAG4D & L4GM & 4Diffusion & SV4D & 4DVD \\ \hline
    Runtime(s) $\downarrow$ & 4200 & 15 & 480 & 565 & \textbf{381(51+330)}  \\
    \bottomrule
  \end{tabular}
  }
\end{table}

\subsection{Runtime Comparison}
Theoretically, as a cascaded diffusion model, 4DVD may require more runtime cost compared with SV4D due to its two-stage coarse-to-fine generation process. However, we find that it is precisely due to the proposed coarse-to-fine generation that 4DVD actually has a runtime advantage over SV4D. We report the runtime cost of different methods on a single A6000 GPU in Tab.~\ref{tab:runtime}. One can see that L4GM realizes the fastest speed as it is not a diffusion model and does not require iterative sampling. Meanwhile, 4DVD is even faster than SV4D. The reason is that, SV4D needs to generate from scratch, so its official code chooses to iterate sampling for 20 steps. In contrast, 4DVD generates in a decoupled and cascade manner. In its first stage, we only aim to produce coarse results with low resolution. Lower resolution allows for faster feedforward, and can be generated within only 10 sampling steps. In the second stage, we have obtained the coarse results as the control hint. Since 4DVD benefits from coarse multi-view priors of the first stage, we also set sampling number to 10, which is enough to complete generation. The runtime cost of both stages are listed in the brackets (first + second) of Tab.~\ref{tab:runtime}.

\subsection{Ablation Study}
\label{subsec:abla}

To validate the effectiveness of our full model, we conduct ablation experiments on the proposed cascaded pipeline. We replace our structure-aware conditional generation branch with existing well-known methods~\cite{controlnet,t2iadapter}, and compare their performance to validate the superiority of our design. Moreover, we replace the proposed MAP with the common cross-attention mechanism (\textit{i.e.}, ``MAP w/ feature'') to demonstrate the effectiveness of our MAP module.

\begin{table}
\centering
  \caption{\textbf{Quantitative results of ablation study.} The best results are highlighted in \textbf{bold}.}
  \label{tab:abla}
  \resizebox{\linewidth}{!}{
  \begin{tabular}{l|ccccc}
    \toprule
    Setting & LPIPS $\downarrow$ & CLIP-S $\uparrow$ & FVD-F $\downarrow$ & FVD-V $\downarrow$ & FVD-Diag $\downarrow$ \\  \hline
    Stage 1 & 0.147 & 0.901 & 668.25 & 422.56 & 605.73  \\
    w/ ControlNet & 0.141 & 0.920 & 558.42  & 407.75  & 535.14   \\
    w/ T2I-adapter & 0.173 & 0.896 & 802.40  & 609.45 &  714.37  \\
    MAP w/ feature & 0.152 & 0.912 & 572.91 & 397.36 & 512.15   \\ \hline
    Full Model & \textbf{0.133} & \textbf{0.927} & \textbf{507.12} & \textbf{314.44} & \textbf{456.01} \\
    \bottomrule
  \end{tabular}
  }
  % \vspace{-0.5em}
\end{table}

\paragraph{Ablation on condition branch.} We replace our structure condition branch with well-known condition-guided methods such as ControlNet~\cite{controlnet} and T2I-adapter~\cite{t2iadapter}. The detailed architectures of these ablation settings can be found in Fig.~\ref{fig:supp_workflow}, where we apply our proposed control branch, T2I-adapter, and ControlNet to the mainstream model to realize coarse-to-fine 4D generation in Fig.~\ref{fig:supp_workflow}(a), (b), (c), respectively. We compare their visual results in Fig.~\ref{fig:supp_abla}. Stage 1 produces coarse results with low visual quality, which mainly aims to provide multi-view layout priors. To generate spatio-temporal content based on these, we apply various architecture to find a feasible injection strategy. Using a T2I-Adapter that only injects features to the layers of the encoder part of mainstream even corrupts the layouts from stage 1, causing performance degradation. Although ControlNet improves quality to a certain extent, this improvement is minor. In contrast, our full model realizes impressive performance gains. Note that all these settings include the proposed MAP module, thus fully demonstrating the superiority of the control branch architecture we developed. In addition, results from stage 1 are sub-optimal and even distorted, such as the hands and feet of the cartoon character. Existing generation methods struggle to rectify these flaws, but our model is capable of correcting them. For a more objective demonstration, quantitative results are given in Tab.~\ref{tab:abla}, which shows our full model works better.

% As shown in Fig.~\ref{fig:supp_workflow}: (a) is our model used in the paper, (b) follows the architecture in T2I-Adapter~\cite{t2iadapter}, and (c) is the ControlNet~\cite{controlnet}.

\paragraph{Ablation on MAP module.} Although the coarse 16-view videos provide layout priors, high-quality production requires more sophisticated content. To fully utilize the input monocular video, we design the MAP attention module to combine its high-quality feature with multi-view layouts. Considering that the coarse results contain 16 views but the reference video only records one viewpoint, we elaborate a novel cross attention to fuse them. Conventional operations \cite{MasaCtrl,FateZero} usually choose features of the current model as \textbf{Q} matrix, and calculate \textbf{KV} matrices based on the external condition. We initially used a similar calculation but found that the trained model tends to introduce obvious visual artifacts as shown in ``MAP w/ feature'' of Fig.~\ref{fig:supp_abla}. Therefore, we take the high-quality reference video as \textbf{Q} matrix and use coarse features to calculate the \textbf{KV} matrices as the conditions in our branch. Experiments demonstrate that this operation works better in our pipeline, bringing better performance. Tab.~\ref{tab:abla} shows that compared with ``MAP w/ feature'', our full model achieves better results. These experiments demonstrate that compared with upscaling multi-view content based on a single view, extending high-quality reference to novel views based on the corresponding layouts is more efficient.

\section{Limitations and Conclusion}

\paragraph{Limitation} We develop 4DVD for dynamic 3D objects generation based on monocular videos, and achieve better results than previous works. However, there are still some challenges to be addressed. The current model may not perform well in complex cases with detailed geometry and complex motions due to the capacity of the network and training data. Nevertheless, 4DVD can serve as a basic model for future research, and the generation of challenging cases can be improved by incorporating more delicate designs and high-quality training data. We believe the same method can also be applied to diffusion transformers~\cite{cogvideo, cogvideox} to address complex situations better and leave it as future work.

%Second, to make 4DVD really useful for 3D developers, some 4D editing functions should be considered in the generation process, especially robust human-in-the-loop ability.

In conclusion, we introduce 4DVD, a cascaded latent video diffusion model designed for multi-view video generation and 4D creation, which could generate temporally consistent and high-resolution videos from a monocular video input. Previous methods attempted to directly model high-dimensional data such as 4D. In contrast, we achieve much more efficient performance by decoupling 4D generation into multi-view layout prediction and structure-aware conditional generation. Benefits from its unprecedented dense view output, the produced results remain high consistency across time and views, while explicit 4D representations, such as 4D Gaussian, can be effectively reconstructed. The model learns temporal consistent multi-view layouts by training simultaneously with 16 views, and utilizes the curated MAP module to achieve high-quality conditional spatio-temporal generation. Extensive experiments demonstrate 4DVD enables more multi-view and temporally consistent 4D generation than existing methods. As an attempt at automatic 4D creation, we believe the proposed cascaded idea provides valuable insight for future research and serves as a basic model for 4D content generation. Moreover, the proposed MAP module enables cross-view consistent conditional multi-view generation. We believe this can also be applied to scene-level generation and leave it as our future work.

\textbf{Acknowledgments}. This work is supported by Guangdong Provincial Key Laboratory of Ultra High Definition Immersive Media Technology (Grant No. 2024B1212010006).

\textbf{Data Availability}. The Objaverse dataset used in this paper is available at \url{https://objaverse.allenai.org/}.

\bibliography{sn-bibliography}

\end{document}